\documentclass[10pt,twocolumn,letterpaper]{article}

\usepackage{iccv}
\usepackage{times}
\usepackage{epsfig}
\usepackage{graphicx}
\usepackage{amsmath}
\usepackage{amssymb}
\usepackage{xcolor}

\usepackage[pagebackref=true,breaklinks=true,letterpaper=true,colorlinks,bookmarks=false]{hyperref}

\iccvfinalcopy 

\def\iccvPaperID{6204} 

\definecolor{todo}{rgb}{7,.0,.3} 

\begin{document}

\title{Guessing Smart: Biased Sampling for Efficient Black-Box Adversarial Attacks}

\author{
	\begin{tabular}[t]{c}
		Thomas Brunner\textsuperscript{1,2}
		\and
		Frederik Diehl\textsuperscript{1,2}
		\and
		Michael Truong Le\textsuperscript{1,2}
		\and
		Alois Knoll\textsuperscript{2}
	\end{tabular} \\
	\begin{tabular}[t]{c}
		\textsuperscript{1} fortiss GmbH
		\and
		\textsuperscript{2} Technical University of Munich
	\end{tabular}\\
	\begin{tabular}[t]{c}
		{\tt\small \{brunner,diehl,truongle\}@fortiss.org}
		\and
		{\tt\small knoll@in.tum.de}
	\end{tabular}
}

\maketitle

\begin{abstract}
We consider adversarial examples for image classification in the black-box decision-based setting. Here, an attacker cannot access confidence scores, but only the final label. Most attacks for this scenario are either unreliable or inefficient. Focusing on the latter, we show that a specific class of attacks, Boundary Attacks, can be reinterpreted as a biased sampling framework that gains efficiency from domain knowledge. We identify three such biases, image frequency, regional masks and surrogate gradients, and evaluate their performance against an ImageNet classifier. We show that the combination of these biases outperforms the state of the art by a wide margin. We also showcase an efficient way to attack the Google Cloud Vision API, where we craft convincing perturbations with just a few hundred queries. Finally, the methods we propose have also been found to work very well against strong defenses: Our targeted attack won second place in the NeurIPS 2018 Adversarial Vision Challenge. 
\end{abstract}


\section{Introduction}

Ever since the term was fist coined, adversarial examples have enjoyed much attention from machine learning researchers. The fact that tiny perturbations can lead otherwise robust-seeming models to misclassify an input could pose a major problem for safety and security. But when discussing adversarial examples, it is often unclear how realistic the scenario of a proposed attack truly is. In this work, we consider a threat setting with the following parameters:

\begin{figure}[htbp]
	
	\centering
	(a) \includegraphics[width=.20\textwidth]{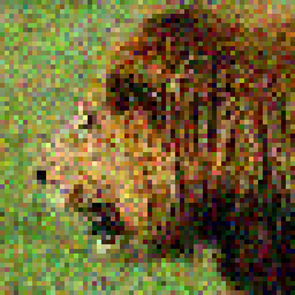}
	\hfill
	(b) \includegraphics[width=.20\textwidth]{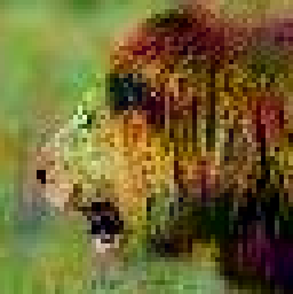}
	
	\ \\
	
	(c) \includegraphics[width=.20\textwidth]{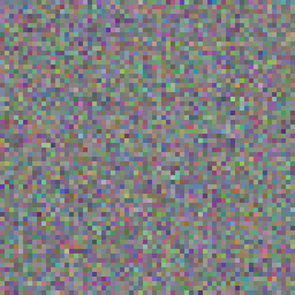}
	\hfill
	(d) \includegraphics[width=.20\textwidth]{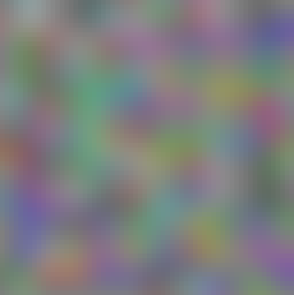}
	
	\caption{(a) and (b): Black-box adversarial examples obtained by random sampling. (c) and (d): isolated perturbation patterns. (c) is sampled from a normal distribution, and (d) is sampled from a distribution of Perlin noise patterns, which is one of the biases we propose. Both (a) and (b) fool the classifier, but (b) can be obtained with fewer samples.}
	\label{fig:figure-perlin}	
\end{figure}



\textbf{Black-box}. 
The black-box setting assumes that an attacker has access only to the input and output of a model. Compared to the white-box setting, where an attacker has complete access to the architecture and parameters of the model, attacks in this setting are significantly harder to conduct: Most state-of-the-art white-box attacks \cite{goodfellow2014, carlini2017towards, madry_towards_2017} rely on gradients that are directly computed from the model parameters, which are not available in the black-box setting. 

\pagebreak

\textbf{Decision-based classification (label-only)}. Depending on the output format of the model, the problem of missing gradients can be circumvented. In a score-based scenario, the model provides real-valued outputs (for example, softmax activations). By applying tiny modifications to the input, an attacker can estimate gradients by observing changes in the output \cite{zoo} and then follow this estimate to generate adversarial examples. The decision-based setting, in contrast, provides only a single discrete result (the top-1 label) on which gradient estimation is very inefficient \cite{ilyas18a}. This form of black-box attack is much more difficult, but also extends the range of possible targets in the real world \cite{brendel17}.

\textbf{Limited queries}. Black-box attacks might not be feasible if they need millions of queries to the model, and possibly multiple hours' time, to be successful. We therefore consider a scenario where the attacker must find a convincing adversarial example in less than 15000 queries.

\textbf{Targeted}. An untargeted attack is considered successful when the classification result is any label other than the original. Depending on the number and semantics of the classes, it can be easy to find a  label that requires little change, but is considered adversarial (\eg Egyptian cat vs Persian cat). A targeted attack, in contrast, needs to produce exactly the specified label. This task is strictly harder than the untargeted attack, further decreasing the probability of success.


In this setting, current state-of-the-art attacks are either unreliable or inefficient. Our contribution is as follows:
\begin{itemize}
	\item We show how a recently proposed method, the Boundary Attack, can be re-framed as a biased sampling framework that gains efficiency from prior beliefs about the target domain.
	\item We discuss three such biases: low-frequency patterns, regional masks and gradients from a surrogate model.
	\item We evaluate the effectiveness of each bias and show that their combination drastically outperforms the previous state of the art in label-only black box attacks. Our source code is publicly available.\footnote{\url{https://github.com/ttbrunner/biased_boundary_attack}}
\end{itemize}

\section{Related Work} \label{sec:related_work}

There currently exist two major schools of attacks in the threat setting we consider:

\subsection{Transfer-based}

It is known that adversarial examples display a high degree of transferability, even between different model architectures \cite{Tramr2017TheSO}. Transfer attacks seek to exploit this by training substitute models that are reasonably similar to the model under attack, and then apply regular white-box attacks to them.

Typically, this is performed by iterative applications of fast gradient-based methods such as the Fast Gradient Sign Method (FGSM) \cite{goodfellow2014} and, more generally, Projected Gradient Descent (PGD) \cite{madry_towards_2017}. The black-box model is used in the forward pass, while the backward pass is performed with the surrogate model \cite{obfuscatedgradients}. In order to maximize the chance of a successful transfer, newer methods use large ensembles of substitute models, and applying adversarial training to the substitute models has been found to increase the probability of finding strong adversarial examples even further \cite{tramer17ens}.

Although these methods currently form the state of the art in label-only black-box attacks \cite{kurakin_nips2017competition}, they have one major weakness: as soon as a defender manages to reduce transferability, direct transfer attacks often run a risk of complete failure, delivering no result even after thousands of iterations. As a result, conducting transfer attacks is a cat-and-mouse game between attacker and defender, where the attacker must go to great lengths to train models that are just as robust as the defender's. Therefore, transfer-based attacks can be very efficient, but also somewhat unreliable.

\subsection{Sampling-based}

Circumventing this problem, sampling-based attacks do not rely on direct transfer and instead try to find adversarial examples by randomly sampling perturbations from the input space.
 
Perhaps the simplest attack consists of sampling a hypersphere around the original image, and drawing more and more samples until an adversarial example is found. Owing to the high dimensionality of the input space, this method is very inefficient and has been dismissed as completely unviable \cite{szegedy13}. While this is not our main focus, we show in Appendix \ref{sec:appendix_avc} that even this crude attack can be accelerated and made competitive in certain scenarios.

Recently, a more efficient attack has been proposed: the Boundary Attack (BA) \cite{brendel17}. This attack is initialized with an input of the desired class, and then takes small steps along the decision boundary to reduce the distance to the original input. Previous works have established that regions which contain adversarial examples often have the shape of a "cone" \cite{tramer17ens}, which can be traversed from start to finish. At each step, the BA employs random sampling to find a sideways direction that leads deeper into this cone. From there, it can then take the next step towards the target.

The BA has been shown to be very powerful, producing adversarial examples that are competitive with even the results of state-of-the-art white-box attacks \cite{brendel17}. However, its weakness is query efficiency: to achieve these results, the attack typically needs to query a model hundreds of thousands of times.


It should be noted that recent research on black-box attacks has largely focused on classifiers that provide confidence scores, which is an easier setting. Nevertheless, many of these methods also use random sampling \cite{zoo,autozoom,ilyas18a}, and the biases we propose could also benefit their approaches. As an aside, Ilyas et al. \cite{ilyas18a} propose a variation of their attack that manages to apply gradient estimation to discrete labels. Although this does fit our setting, we find it to be much less efficient than BA variants (see Section \ref{sec:eval}).

Clearly, sampling-based attacks are very flexible but often too inefficient for practical use. Barring pure random guessing, the BA is the simplest attack for our setting. We therefore choose to focus on this method, and show how it can benefit from the biases we propose.

\begin{figure*}[t]
	\centering
	(a) \includegraphics[width=.3\textwidth]{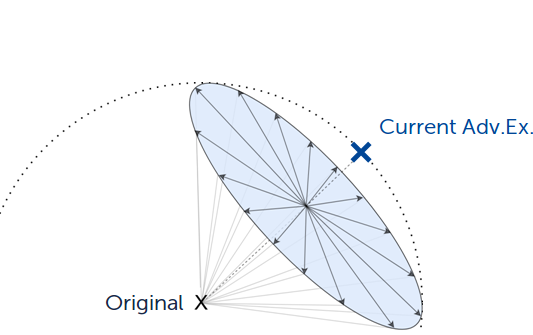}
	\hfill
	(b) \includegraphics[width=.3\textwidth]{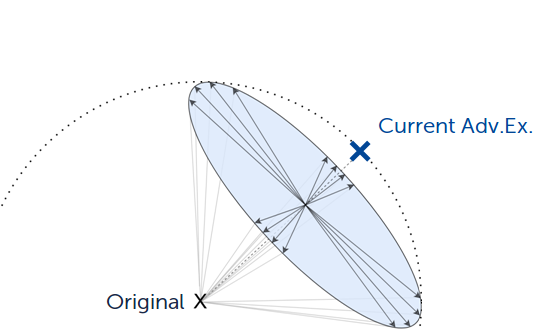}
	\hfill
	(c) \includegraphics[width=.3\textwidth]{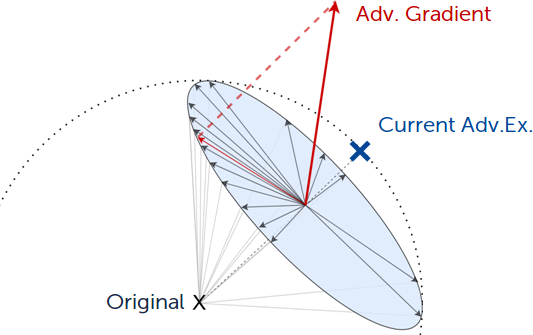}
	
	\caption{Sampling directions for the orthogonal step. (a) Boundary Attack: uniformly distributed along the surface of the hypersphere. (b) BA with Perlin bias: higher sample density in the direction of low-frequency perturbations. (c) BA with surrogate gradient bias: samples further concentrate towards the direction of the projected gradient.}
	\label{fig:figure-bba}	
\end{figure*}
\pagebreak

\section{Biased Boundary Attacks}

The Boundary Attack, like most sampling-based attacks, draws perturbation candidates from a multidimensional normal distribution. This means that it performs unbiased sampling, perturbing each input feature independently of the others. While this is very flexible, it is also extremely inefficient when used against a robust model.

Consider the distribution of natural images: adjacent pixels are typically not independent of each other, but often have similar colors. This alone is a strong indicator that drawing perturbations from i.i.d random variables will lead to adversarial examples that are clearly out of distribution for natural image datasets. This, of course, renders them vulnerable to detection and filtering - robust models have become increasingly resilient against such patterns \cite{kurakin_nips2017competition}.

Therefore, it seems only logical to constrain the search space to perturbations that we believe to have a higher chance of success, or to bias the distribution so that the probability of sampling them increases.

We outline three such biases for the domain of image classification, discuss their motivation and show how to integrate them into the sampling procedure of the Boundary Attack.

\subsection{Low-frequency perturbations}


 When one looks at typical adversarial examples, it quickly becomes apparent that most existing methods yield perturbations with high image frequency. But high-frequency patterns have a significant problem: they are easily identified and separated from the original image signal, and are often dampened by spatial transforms. Indeed, most of the winning defenses in the NeurIPS 2017 Adversarial Attacks and Defences Competition were based on denoising \cite{dunet}, simple median filters \cite{kurakin_nips2017competition}, and random transforms \cite{xie2018mitigating}. In other words: state-of-the art defenses are designed to filter high-frequency noise.

At the same time, we know that it is possible to synthesize "robust" adversarial examples which are not easily filtered in this way. Compare Athalye et al. \cite{Athalye2018SynthesizingRA}: Their robust perturbations are largely invariant to filters and transforms, and -- interestingly enough -- at first glance seem to contain very little high-frequency noise.

Inspired by this observation, we hypothesize that image frequency alone could be a key factor in robustness of adversarial perturbations. If true, then simply limiting perturbations to the low-frequency domain should increase the success chance of an attack, while incurring no extra cost.

\textbf{Perlin Noise patterns}. A straightforward way to generate parametrized, low-frequency patterns, is to use Perlin Noise \cite{Perlin85}. Originally intended as a procedural texture generator for computer graphics, this function creates low-frequency noise patterns with a reasonably "natural" look. One such pattern can be seen in Figure \ref{fig:figure-perlin}d. But how can we use it to create a prior for the Boundary Attack?

Let $k$ be the dimensionality of the input space. The original Boundary Attack (Figure \ref{fig:figure-bba}a) works by applying an orthogonal perturbation $\eta^k$ along the surface of a hypersphere around the original image, in the hope of moving deeper into an adversarial region. From there, a step is taken towards the original image. In the default configuration, candidates for $\eta^k$ are generated from samples $s \sim \mathcal{N}(0, 1)^k$, which are projected orthogonally to the source direction and normalized to the desired step size. This leads to the directions being uniformly distributed along the hypersphere.

To introduce a low-frequency prior into the Boundary Attack, we instead sample from a distribution of Perlin noise patterns (Figure \ref{fig:figure-bba}b). Perlin noise is typically parametrized with a permutation vector $v$ of size 256, which we randomly shuffle on every call. Effectively, this allows us to sample two-dimensional noise patterns $s \sim Perlin_{h,w}(v)$, where $h$ and $w$ are the image dimensions (and $h\cdot w=k$). As a result, the samples are now strongly concentrated in low-frequency directions. 

Our experiments in Section \ref{sec:eval} show that this greatly improves the efficiency of the attack. Therefore, we reason that the distribution of Perlin noise patterns contains a higher concentration of adversarial directions than the normal distribution.


\begin{figure}[htbp]
	\centering
	(a) \includegraphics[width=.20\textwidth]{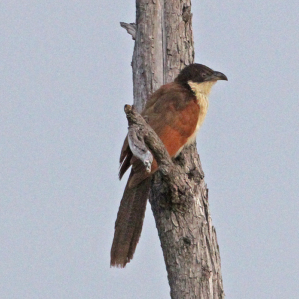} \hfill 
	(b) \includegraphics[width=.20\textwidth]{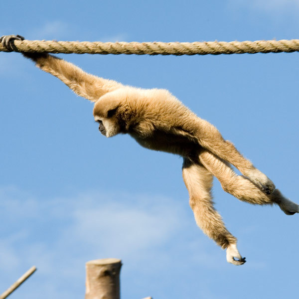}  
	\\ \ \\
	(c) \includegraphics[width=.20\textwidth]{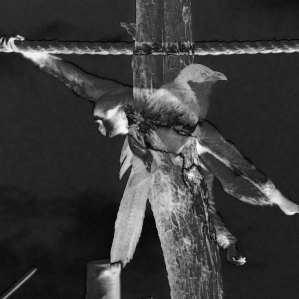} \hfill
	(d) \includegraphics[width=.20\textwidth]{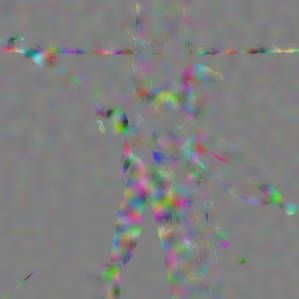}

	\caption{Masking based on per-pixel difference. (a) shows the original image, (b) an image of the target class. (c) is the mask, and (d) is a perturbation to which the mask has been applied. The perturbation concentrates on the central region, as the background is already quite similar between the images. }
	\label{fig:mask}	
\end{figure}

\subsection{Regional masking}

Currently, perturbations are evenly applied across the entire image. No matter if low or high frequency, the orthogonal step of the Boundary Attack perturbs all pixels nearly equally (when averaged over a large number of samples). This seems to be a waste -- could the attack benefit from limiting the perturbation to specific regions?

The Boundary Attack is an interpolation from an image of the target class towards the image under attack. In some regions, these images might already be quite similar, while being very different in others. Intuitively, we would want an attack to take larger steps in those regions where the difference is high. We are also reluctant to perturb regions that are already similar, as any such distortion will have to be undone in a later step. 

It turns out that this is an ideal way to reduce the search space. We can simply create an image mask from the per-pixel difference of the adversarial and original image (see Figure \ref{fig:mask}):
\begin{equation}
M = |X_{adv} - X_{orig}|
\end{equation}

At each step, we recalculate this mask based on the current position and apply it element-wise to the previously sampled orthogonal perturbation $\eta^k$:

\begin{equation}
\eta^k_{biased} = M\odot\eta^k; \  \eta^k_{biased} = \frac{\eta^k_{biased}}{\|\eta^k_{biased}\|}
\end{equation}

In this way, the distortion of those pixels that have a high difference is amplified and that of similar pixels dampened, while the magnitude of the perturbation vector stays the same. 

This reduces the search space of the attack and therefore increases its efficiency -- if one assumes our intuition about regional masking to be correct. We implement this masking strategy as a proof of concept and our evaluation in Section \ref{sec:eval} shows that it indeed improves efficiency by a significant amount.

\textbf{Other masks.} An attacker might wish to engineer masks from other knowledge they possess about the image contents. For example, it could be worthwhile to concentrate the perturbation on the most salient features of the target class, reducing the search space to only the most vital dimensions. Such "focused" black-box perturbations could hold much promise, and we aim to investigate them in the future.

\subsection{Gradients from surrogate models}


What other source of information contains strong hints about directions that are likely to point to an adversarial region? Naturally -- gradients from surrogate models. Transfer attacks have been shown to be extremely powerful (albeit brittle) \cite{tramer17ens}, so it should be useful to exploit surrogates whenever they are available.

Arguably, the main weakness of transfer attacks is that they fail when the decision boundary of the surrogate model does not closely match the defender's. However, even when this is the case, the boundary may still be reasonably nearby. Based on this intuition, some approaches extend gradient-based attacks with limited regional sampling \cite{obfuscatedgradients}. Here, we do exactly the opposite and extend a sampling-based attack with limited gradient information. This has the significant advantage that, in the case of low transferability, our method merely experiences a slowdown where typical transfer attacks completely fail.

Our method works as follows:

\begin{itemize}
	\item An adversarial gradient from a surrogate model is calculated. Since the current position is already adversarial, it can be helpful to move a small distance towards the original image first, making sure to calculate the gradient from inside a non-adversarial region.
	
	\item The gradient usually points away from the original image, therefore we project it orthogonally to the source direction, as shown in Figure \ref{fig:figure-bba}c.
	
	\item This projection is on the same hyperplane as the candidates for the orthogonal step. We can now bias the candidate perturbations toward the projected gradient by any method of our choosing. Provided all vectors are normalized, we opt for simple addition:
	
	\begin{equation}
	\eta^k_{biased} = (1-w)\cdot\eta^k + w\cdot\eta^k_{PG}
	\end{equation}
	
	\item $w$ controls the strength of the bias and is a hyperparameter that should be tuned according to the performance of our substitute model. High values for $w$ should be used when transferability is high, and vice versa. Were we to choose the maximum value, $w$ = 1, then the orthogonal step would be equivalent to an iteration of the PGD attack. In our experience, $w\leq 0.5$ generally leads to good performance.
	
\end{itemize}

As a result, samples concentrate in the vicinity of the projected gradient, but still cover the rest of the search space (albeit with lower resolution). In this way, substitute models are purely optional to our attack instead of forming the central part. It should be noted though that at least \textit{some} measure of transferability should exist. Otherwise, the gradient will point in a bogus direction and using a high value for $w$ would reduce efficiency instead of improving it. 

For the time being, this does not pose a major problem. To the best of our knowledge, no strategies exist that successfully eliminate transferability altogether. As we go on to show in Section \ref{sec:eval}, even surrogate models that are too weak for direct transfer attacks can be used in our framework.

\subsection{Concurrent work}

Our work is concurrent with Ilyas et al. \cite{Ilyas2018PriorCB}, who introduce a bandit optimization framework that incorporates prior information in order to increase query efficiency. While their approach differs from ours, it is motivated by the same intuition - domain knowledge can be used to speed up optimization. We note that the data-dependent prior they propose is essentially a low-frequency bias not unlike our own. They also introduce a time-dependent prior, which could benefit our work in the future.

Low-frequency perturbations have also recently been described by Guo et al. \cite{guo2018low}. They decompose random perturbations with the Discrete Cosine Transform, and then remove high frequencies from the spectrum. We expect these patterns to be very similar to those produced by our Perlin bias.

\section{Evaluation} 	\label{sec:eval}	

We evaluate our approach against an ImageNet classifier and perform an ablation study to determine the effectiveness of each bias. We also compare our results to a range of recently proposed black-box attacks. Finally, we mount an attack against the Google Cloud Vision API and show that our approach can be efficiently deployed against real-world commercial systems. 

Appendix \ref{sec:appendix_images} shows a range of interesting examples produced by our attacks, Appendix \ref{sec:appendix_hyper} contains a listing of hyperparameters and Appendix \ref{sec:appendix_avc} describes our winning submission to the NeurIPS 2018 Adversarial Vision Challenge in detail.

\begin{figure}[htb]
	\centering
	(a) \includegraphics[width=.205\textwidth]{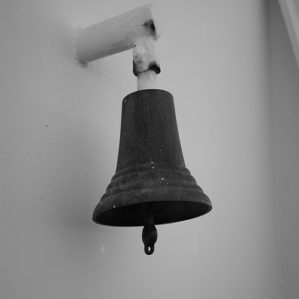} \ 
	(b) \includegraphics[width=.205\textwidth]{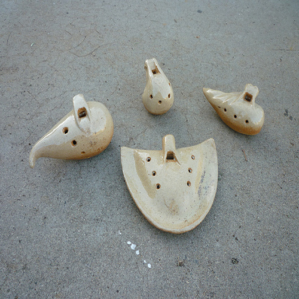} 
	\\
	\ 
	\\
	
	(c) \includegraphics[width=.205\textwidth]{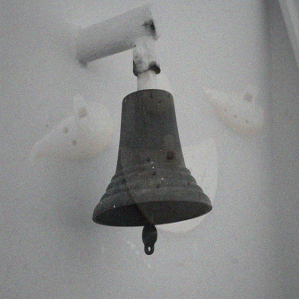} \ 
	(d) \includegraphics[width=.205\textwidth]{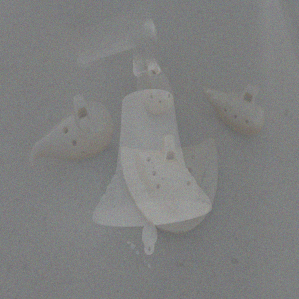} 
	\\
	\ 
	\\
	
	(e) \includegraphics[width=.205\textwidth]{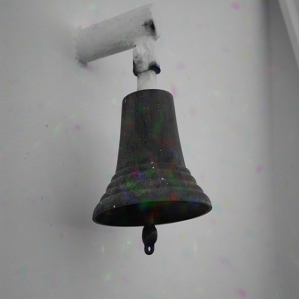} \ 
	(f) \includegraphics[width=.205\textwidth]{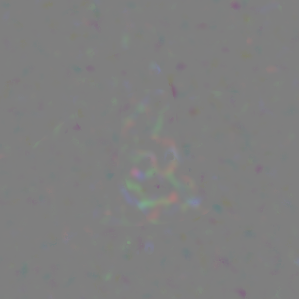} 
	
	\caption{Targeted attack on ImageNet using 15000 queries. (a) shows the original image (bell) and (b) an image of the target class (ocarina). (c) is an adversarial example generated by the original Boundary Attack ($d_{\ell^2}=18.52$), with (d) showing the difference to the original image. (e) and (f) show the biased Boundary Attack with all biases enabled ($d_{\ell^2}=4.78$)}
	\label{fig:bell-bba}	
\end{figure}

\begin{table*}[htp]

	\vskip 0.1in
	\begin{center}
		\begin{small}
			\begin{sc}
				\begin{tabular}{ccc|rrrrrr|c}
					\multicolumn{3}{c|}{Active biases} & \multicolumn{6}{c|}{Success rate vs number of queries} & Median queries \\
					Perlin & Mask & Surrogate  & 500 & 1000 & 2500 & 5000 & 10000 & 15000 & until success \\
					\hline
					no &no& no & 0.01 & 0.03 & 0.10 & 0.17	& 0.25 & 0.33 & - \\					
					
					yes &no& no & 0.03 & 0.08 & 0.18 & 0.31 & 0.47 & 0.57 & 10826	\\					
					
					no & yes & no & 0.03 & 0.04 & 0.13 & 0.22	& 0.35 & 0.44 & - \\					
					
					yes & yes & no & 0.04 & 0.09 & 0.25 & 0.46	& \textbf{0.72} & 0.80 & 5485 \\
					no &no&yes& 0.02 & 0.06 & 0.10 & 0.19	& 0.29 & 0.38 & - \\
					
					yes &no&yes& \textbf{0.05} & 0.08 & 0.17 & 0.32 & 0.50 & 0.60 & 10277	\\
					no & yes &yes& 0.03 & 0.06 & 0.15 & 0.26	& 0.38 & 0.49 & - \\	
					
					yes & yes &yes& 0.04 & \textbf{0.10} & \textbf{0.29} & \textbf{0.48}	& 0.69 & \textbf{0.85} & \textbf{5432} \\
					
				\end{tabular}
			\end{sc}
		\end{small}
	\end{center}
	\caption{Ablation study of biases on ImageNet (targeted attack). The Perlin bias has the strongest effect, followed by the mask bias and finally the surrogate bias. Each bias improves efficiency on its own, and the combination of all biases delivers the strongest performance. \textit{Note: we only report the median for success rates over 0.5.}}
		\label{tab_targ_bias_comp}
	\vskip -0.1in
\end{table*}

\begin{table*}[htp]
	\vskip 0.1in
	\begin{center}
		\begin{small}
			\begin{sc}
				\begin{tabular}{l|cccccc}
					\ & \multicolumn{6}{c}{Success rate vs number of queries} \\
					\ Method & 500 & 1000 & 2500 & 5000 & 10000 & 15000 \\
					\hline
					Ilyas et al. \cite{ilyas18a} (label-only) & 0.00 & 0.00 & 0.00 & 0.00 & 0.00   & 0.00 \\
					Cheng et al. \cite{chengHardlabel} & 0.04 & 0.04 & 0.04 & 0.04 & 0.07 & 0.07 	\\
					Madry et al. \cite{madry_towards_2017} (PGD transfer attack) & \textbf{0.07} & 0.08 & 0.08 & 0.09	& 0.09 & 0.09 \\					
					Brendel et al. \cite{brendel17} (unbiased Boundary Attack) & 0.01 & 0.03 & 0.10 & 0.17	& 0.25 & 0.33 \\
					\textbf{Ours (Biased Boundary Attack)} & 0.04 & \textbf{0.10} & \textbf{0.29} & \textbf{0.48}	& \textbf{0.69} & \textbf{0.85} \\
					
					
				\end{tabular}
			\end{sc}
		\end{small}
	\end{center}
	\caption{Comparison with recently proposed label-only attacks. We use the original code provided by the respective authors and run all methods with the same data and targets. The biased Boundary Attack (same as in Table \ref{tab_targ_bias_comp}) outperforms all other methods. Note that in the case of Ilyas et al.\cite{ilyas18a}, the number of required queries is so large that the attack never achieves success in the range we consider.}	
		\label{tab_imagnet_targ}
\end{table*}

\subsection{ImageNet}
\label{sec:eval-imagenet}

ImageNet consists of images with 299x299 color pixels and has 1000 classes. We run our attacks against a pre-trained InceptionV3 network \cite{inceptionv3}, which achieves 78\% top-1 accuracy. 

\textbf{Evaluation}. We create an evaluation set by randomly selecting 1000 images from the ImageNet validation set while fixing a random target label for each image. We then proceed to run each attack for up to 15000 queries and measure the \textit{success rate} over all examples.

\textbf{Success criteria}. The Boundary Attack always starts with an image of the adversarial class, therefore success must be characterized by low distance to the original image. For this, we define a threshold on the $\ell^2$-norm of the adversarial perturbation and register success when the distance is below. Since some of the methods we compare \cite{ilyas18a} use $\ell^\infty$ distance exclusively, we set the $\ell^2$-threshold to 25.89. This corresponds to a worst-case $\ell^\infty$-distortion of 0.05 if one assumes all pixels to be maximally perturbed.

\textbf{Initialization}. We search the ImageNet validation set for an image of the target class, and pick the one that is closest to the one being attacked. From there, we perform a binary line search to find the decision boundary. This is typically done in less than 10 queries. We run all attacks with the same starting points.

\textbf{Surrogate model}. We use a pre-trained  Inception-Resnet-v2 model \cite{Szegedy2016Inceptionv4IA} for the gradient bias. This model is not adversarially trained and, as Table \ref{tab_imagnet_targ} shows, performs poorly in a PGD transfer attack. We intentionally use this model to demonstrate that our approach can effectively use pre-trained surrogates without any need for modification.

\textbf{Hyperparameters}. See Appendix \ref{sec:appendix_hyper}.

\subsubsection{Ablation study of biases}

We first evaluate all three biases and their combinations. Table \ref{tab_targ_bias_comp} shows the result: it is apparent that each of the biases increases the efficiency of the attack. The largest boost is obtained by the Perlin bias, followed by the mask bias, and finally the surrogate gradient bias.

The latter has a rather small effect, which is probably due to the fact that our surrogate model is too weak. But still, we are able to use what little transferability there is instead of slowing down (or failing like a transfer attack would). At this point, it would be interesting to see whether our method could profit even more from better surrogates. We consider this a direction for future work.

It is also noteworthy that all biases can be combined, and that they do not interfere with each other. When all three biases are active, we reach 85\% success after only 15000 queries. Figure \ref{fig:bell-bba} shows an example of this drastic improvement. This is our strongest attack, which we now compare against other state-of-the-art methods.

\subsubsection{Comparison to state of the art}
\label{sec:eval-comparison}

We go on to benchmark our method against a range of recently proposed attacks for our setting. We use publicly available code, together with the hyperparameters recommended by the authors. We modify the implementations to use our evaluation set, therefore all attacks are run on the same 1000 images and use the same target labels as well as starting points (where applicable). Table \ref{tab_imagnet_targ} shows the results. 

\textbf{Ilyas et al. \cite{ilyas18a}} propose a label-only version of their gradient estimation attack. It works in our setting, but at greatly reduced efficiency. We note that it does not produce an adversarial example within 15000 queries and that no run succeeds before 276000 queries, with a median of 2.48 million queries required. This is in line with their published results. Their other attacks use confidence scores, and therefore require a setting that is considerably easier. Even then, our median number of queries is decisively lower than the one reported by them (5432 versus 11550).

Similarly, \textbf{Cheng et al. \cite{chengHardlabel}} re-frame the setting as a real-valued optimization problem. In general, they report higher efficiency than the Boundary Attack, which is not confirmed in the setting of our experiment. We used their publicly available source code with recommended hyperparameters.

Finally, we perform an iterative PGD transfer attack as described by \textbf{Madry et al.} \cite{madry_towards_2017}. Our results show that its performance is hit-and-miss: when a transfer succeeds it does so very early, but in most cases it never succeeds. Clearly this attack requires a stronger surrogate model.

Interestingly enough, the performance we obtain for the original Boundary Attack (without biases) seems higher than that observed in previous work \cite{brendel17,chengHardlabel}. This may be due to our initialization method or our choice of hyperparameters, which we list in Appendix \ref{sec:appendix_hyper}. In any case, our evaluation shows that the biased Boundary Attack decidedly outperforms all other attacks in a label-only setting.

\subsection{Google Cloud Vision API}
\label{google_attack}

To show that our method is effective even against black boxes with unknown labels, we conduct an attack against the Google Cloud Vision API. This is significantly harder than attacking ImageNet, since the exact classes are unknown. However, we also note that Google Cloud Vision has a very high number of near-redundant class labels. Do we really need to focus on one label alone?

\textbf{Free-form attacks}. We have argued earlier that untargeted attacks with many redundant classes are not truly adversarial. However, the opposite is also true: to achieve an adversarial effect, it is not always necessary to target one specific class. Rather, the same effect could be achieved by targeting a group of classes -- if we want to label a dog as a cat, we can take the union of all cat breeds to achieve the desired effect. 

To be more precise, we can formulate \textit{any} adversarial criterion, as long as it is a boolean function of the model output. Decision-based attacks make this very simple: if we consider this function to be part of the black box, we can simply treat it like any other model and run our attack on its output.

\subsubsection{Turning a person into a bear} 

Consider, for example, a targeted attack to turn a person into a bear. Instead of using the label "bear", we perform a string comparison on the top-1 label and check for the occurrence of "bear". This extends the attack to labels like "grizzly bear", "brown bear", etc. and keeps it from getting stuck whenever one of these labels appears in top-1 position. We also add another condition for good measure: the words "face", "facial expression", "skin", "person" must not appear in any of the output labels.

\begin{figure}[tbp]
	\centering
	\includegraphics[width=.45\textwidth]{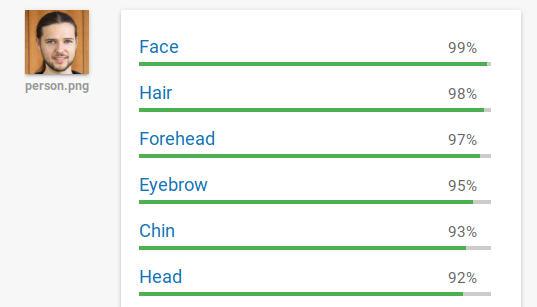} \ 
	\ \\
	\ \\
	\includegraphics[width=.45\textwidth]{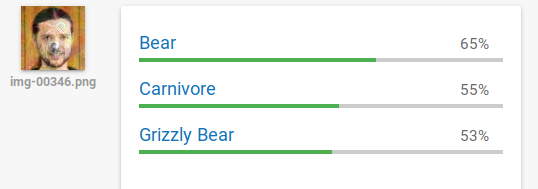} \ 
	\caption{Adversarial image (target "bear"), classified by Google Cloud Vision after 346 queries. Confidence scores are displayed, but not used by the attack. No label hinting at a person is left (the prediction vector contains only 3 labels). }
	\label{fig:googleadv}	
\end{figure}

Figure \ref{fig:googleadv} shows the result: after only 346 iterations, our attack produces a perturbation that is still visible, but small enough to fool an unsuspecting person. 

\begin{figure}[htbp]
	\centering
	\includegraphics[width=.45\textwidth]{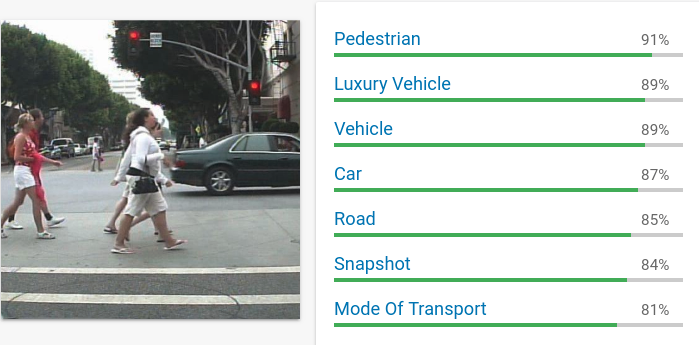} \ 
	\ \\
	\ \\	\includegraphics[width=.45\textwidth]{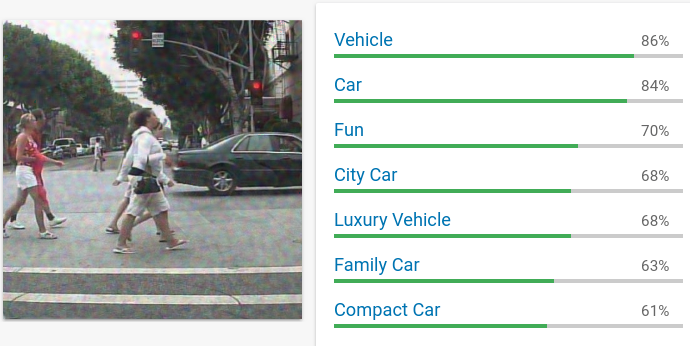} \ 
	\caption{Adversarial image (pedestrians have been removed), classified by Google Cloud Vision after exactly 1000 queries.  Confidence scores are displayed, but not used by the attack. Perhaps interestingly, our adversarial pattern is classified as "Fun".}
	\label{fig:googlepedadv}	
\end{figure}

\subsubsection{Making pedestrians disappear}
The adversarial criterion can also be formulated as a top-k untargeted attack on multiple labels. Consider a potentially safety-critical scenario, where the goal is to make the model oblivious to pedestrians. For this, we simply formulate the condition so that the string "pedestrian" does not appear in the prediction vector, and that related labels such as "person, walking, head, clothes" are also absent.

We obtain Figure \ref{fig:googlepedadv} after exactly 1000 queries. Again, the perturbation is already small enough to fool an unsuspecting observer. An attack with such a low number of queries can be performed on virtually any device -- even mobile -- in a matter of minutes, and is only limited by the latency of the API under attack.


\section{Conclusion}

We have shown that decision-based black-box attacks can be greatly sped up with prior knowledge. The Boundary Attack can be interpreted as a biased sampling framework where one merely needs to modify the distribution from which samples are drawn.

Within this framework, we have proposed three priors that are partially motivated by intuitions about the nature of image classification, and partially by a desire to connect research directions in the field of black-box attacks. Consider the surrogate gradient bias: by itself, it does not yield a substantial improvement. However, the observation that we are able to draw even a small benefit from surrogates that otherwise show near-zero transferability seems very promising for future work. We aim to study this effect of partial transferability in more detail and hope to uncover some of its underlying properties.

And it does not end here - we have discussed only three priors for biased sampling, but there is much more domain knowledge that has not yet found its way into adversarial attacks. Other perturbation patterns, spatial transforms, adversarial blending strategies, or even intuitions about semantic features of the target class could all be integrated in a similar fashion.

With the biased Boundary Attack, we have outlined a basic framework into which a broad range of knowledge can be incorporated. Our implementation significantly outperforms the previous state of the art in black-box label-only attacks, which is one of the most difficult settings currently considered. Our methods can be used to craft convincing results after very few iterations, and the threat of black-box adversarial examples becomes more realistic than ever before.

{\small
\bibliographystyle{ieee}
\bibliography{egbib}
}

\onecolumn
\appendix
\section{Evaluation images}
\label{sec:appendix_images}

\begin{figure*}[htbp!]
	\vskip -0.2in
	\begin{center}
		\begin{tabular}{c c c c}
			
			
			\textbf{No bias} & \textbf{Surrogate bias only} & \textbf{Mask bias only} & \textbf{Mask+Surrogate} \\
			$d_{\ell^2} = 27.10$ & $d_{\ell^2} = 21.73$ & $d_{\ell^2} = 13.19$ & $d_{\ell^2} = 12.40$ \\

			\includegraphics[width=.195\textwidth]{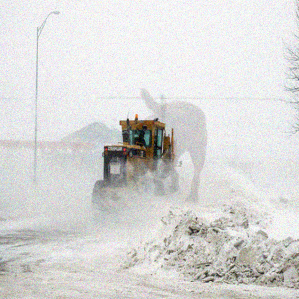} & 
			\includegraphics[width=.195\textwidth]{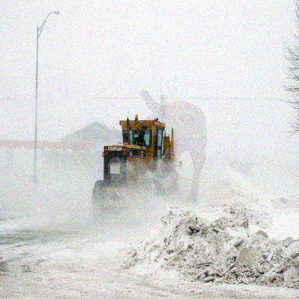} & 
			\includegraphics[width=.195\textwidth]{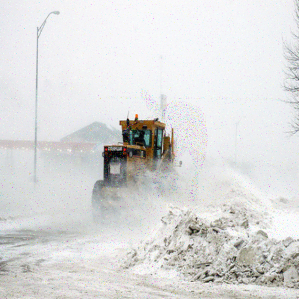} &
			\includegraphics[width=.195\textwidth]{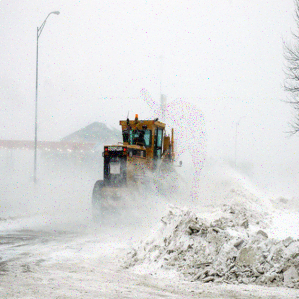} \\

			\includegraphics[width=.195\textwidth]{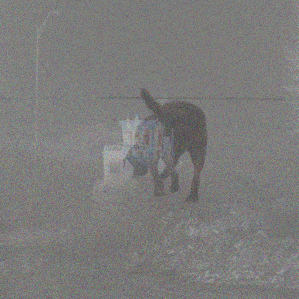} & 
			\includegraphics[width=.195\textwidth]{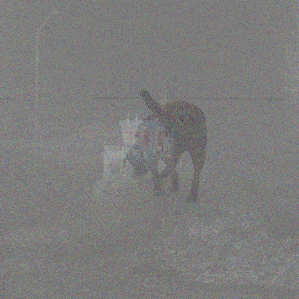} & 
			\includegraphics[width=.195\textwidth]{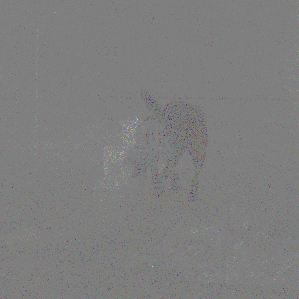} &
			\includegraphics[width=.195\textwidth]{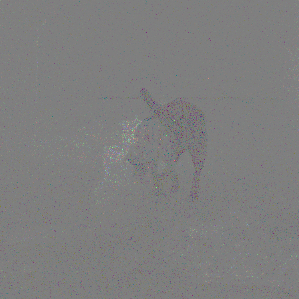} \\

			\\
			\textbf{Perlin bias only} & \textbf{Perlin+Surrogate} & \textbf{Perlin+Mask} & \textbf{Perlin+Mask+Surrogate} \\
			$d_{\ell^2} = 20.40$ & $d_{\ell^2} = 17.89$ & $d_{\ell^2} = 8.79$ & $d_{\ell^2} = 8.28$ \\
			\includegraphics[width=.195\textwidth]{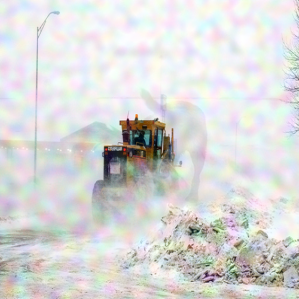} & 
			\includegraphics[width=.195\textwidth]{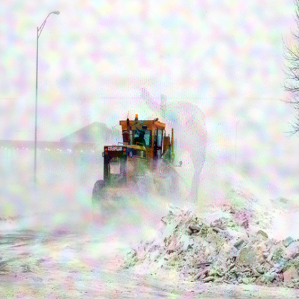} & 
			\includegraphics[width=.195\textwidth]{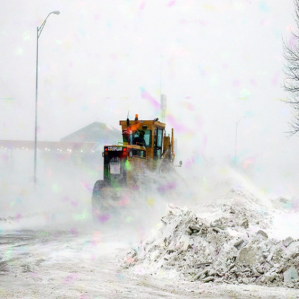} & 
			\includegraphics[width=.195\textwidth]{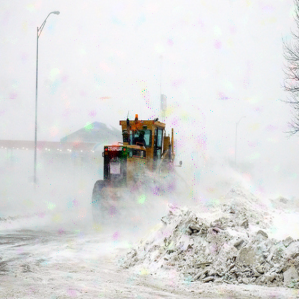} \\
			
			\includegraphics[width=.195\textwidth]{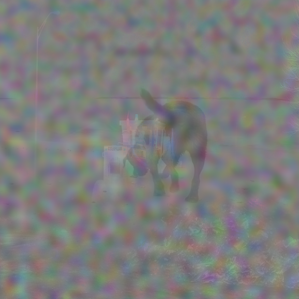} & 
			\includegraphics[width=.195\textwidth]{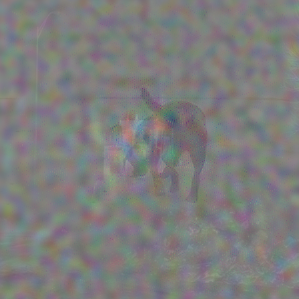} & 
			\includegraphics[width=.195\textwidth]{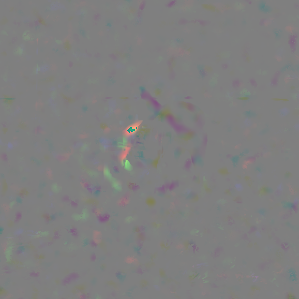} & 
			\includegraphics[width=.195\textwidth]{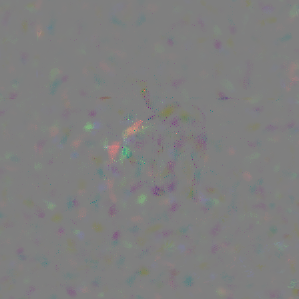} \\
			
			\\
			\textbf{Original image} & \textbf{Starting point} & & \\
			\includegraphics[width=.195\textwidth]{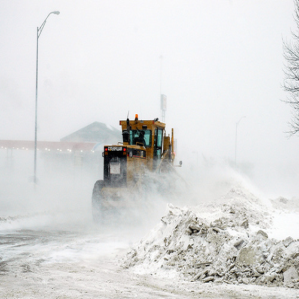} & 
			\includegraphics[width=.195\textwidth]{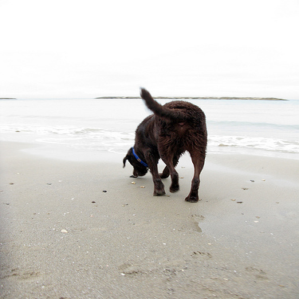} & 
			& 
			\\
			
		\end{tabular}	
		
		\label{fig-extra-plow}
	\end{center}
	\vskip -0.1in
	\caption{Targeted adversarial examples on ImageNet, obtained with different biases after 15000 iterations. The original class is "snowplow" -- all images are classified as the target "Chesapeake Bay retriever". The mask bias is especially effective, as start and original image have similar backgrounds. See \url{https://github.com/ttbrunner/biased_boundary_attack} for an animated version.}
\end{figure*}

\begin{figure*}[htbp!]
	\vskip -0.1in
	\begin{center}
		\begin{tabular}{c c c c}
			
			
			\textbf{No bias} & \textbf{Surrogate bias only} & \textbf{Mask bias only} & \textbf{Mask+Surrogate} \\
			$d_{\ell^2} = 42.36$ & $d_{\ell^2} = 40.51$ & $d_{\ell^2} = 42.69$ & $d_{\ell^2} = 41.69$ \\

			\includegraphics[width=.195\textwidth]{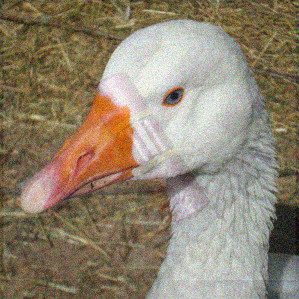} & 
			\includegraphics[width=.195\textwidth]{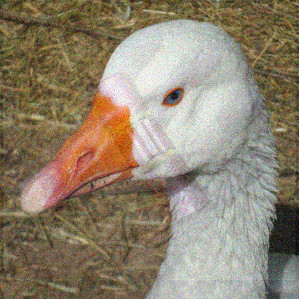} & 
			\includegraphics[width=.195\textwidth]{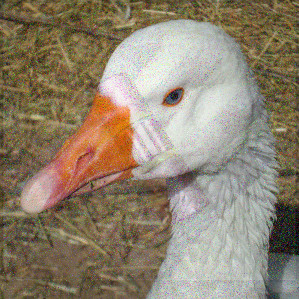} &
			\includegraphics[width=.195\textwidth]{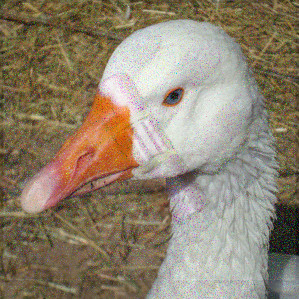} \\

			\includegraphics[width=.195\textwidth]{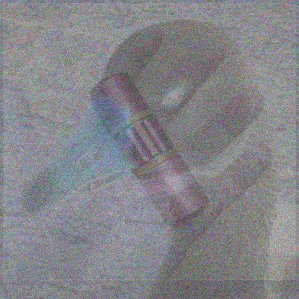} & 
			\includegraphics[width=.195\textwidth]{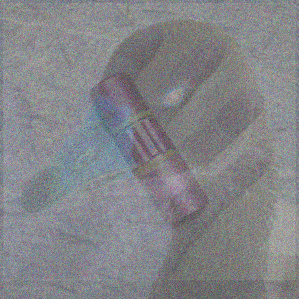} & 
			\includegraphics[width=.195\textwidth]{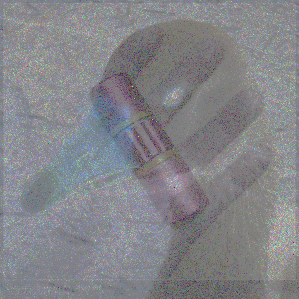} &
			\includegraphics[width=.195\textwidth]{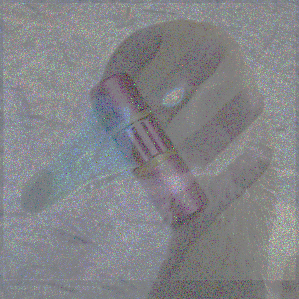} \\

			\\
			\textbf{Perlin bias only} & \textbf{Perlin+Surrogate} & \textbf{Perlin+Mask} & \textbf{Perlin+Mask+Surrogate} \\
			$d_{\ell^2} = 34.92$ & $d_{\ell^2} = 35.19$ & $d_{\ell^2} = 12.37$ & $d_{\ell^2} = 10.54$ \\
			\includegraphics[width=.195\textwidth]{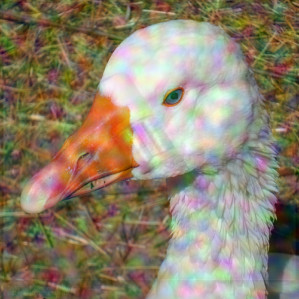} & 
			\includegraphics[width=.195\textwidth]{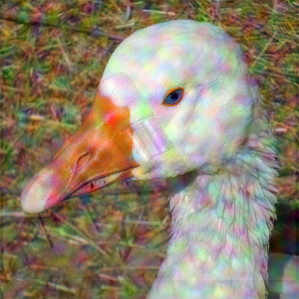} & 
			\includegraphics[width=.195\textwidth]{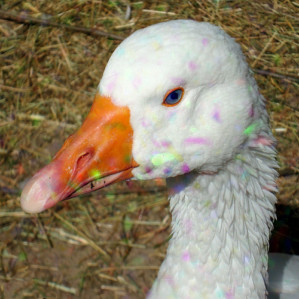} & 
			\includegraphics[width=.195\textwidth]{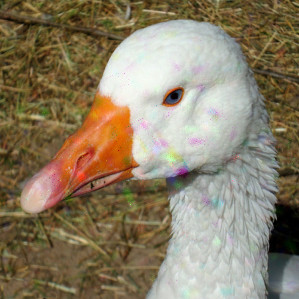} \\
			
			\includegraphics[width=.195\textwidth]{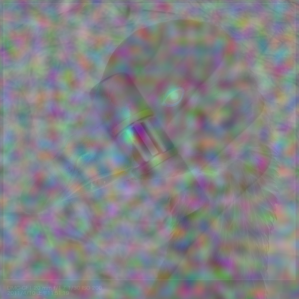} & 
			\includegraphics[width=.195\textwidth]{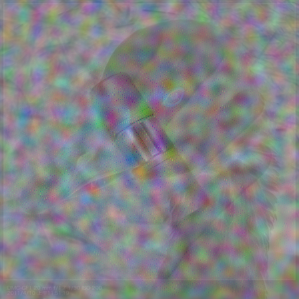} & 
			\includegraphics[width=.195\textwidth]{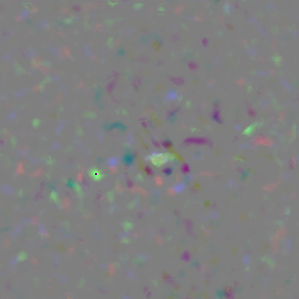} & 
			\includegraphics[width=.195\textwidth]{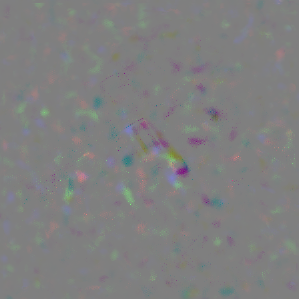} \\
			
			\\
			\textbf{Original image} & \textbf{Starting point} & & \\
			\includegraphics[width=.195\textwidth]{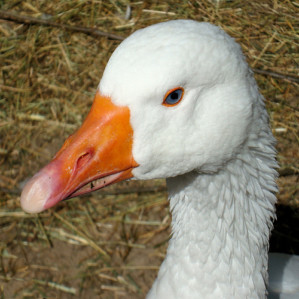} & 
			\includegraphics[width=.195\textwidth]{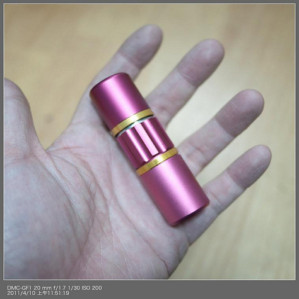} & 
			& 
			\\
			
		\end{tabular}	
		
		\label{fig-extra-goose}
	\end{center}
	\vskip -0.1in
	\caption{Targeted adversarial examples on ImageNet, obtained with different biases after 15000 iterations. The original class is "goose" -- all images are classified as the target "lipstick". In the case of this image, not every bias comes with an improvement: mask+surrogate is even slightly worse than surrogate only. However, when all biases are combined the result is still significantly better. See \url{https://github.com/ttbrunner/biased_boundary_attack} for an animated version.}
\end{figure*}

\begin{figure*}[htbp!]
	\vskip -0.1in
	\begin{center}
		\begin{tabular}{c c c c}
			
			
			\textbf{No bias} & \textbf{Surrogate bias only} & \textbf{Mask bias only} & \textbf{Mask+Surrogate} \\
			$d_{\ell^2} = 47.36$ & $d_{\ell^2} = 57.74$ & $d_{\ell^2} = 40.10$ & $d_{\ell^2} = 41.84$ \\

			\includegraphics[width=.195\textwidth]{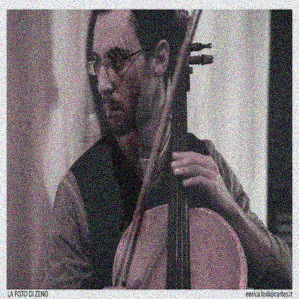} & 
			\includegraphics[width=.195\textwidth]{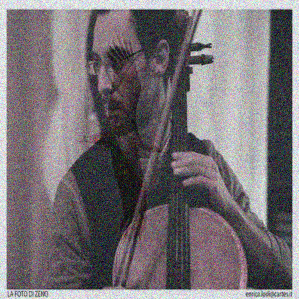} & 
			\includegraphics[width=.195\textwidth]{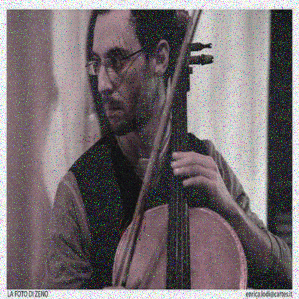} &
			\includegraphics[width=.195\textwidth]{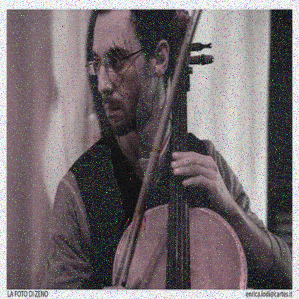} \\

			\includegraphics[width=.195\textwidth]{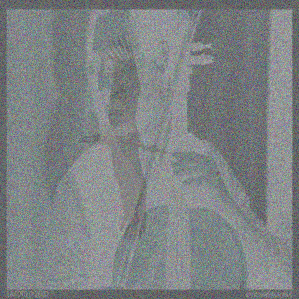} & 
			\includegraphics[width=.195\textwidth]{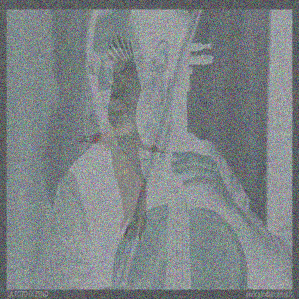} & 
			\includegraphics[width=.195\textwidth]{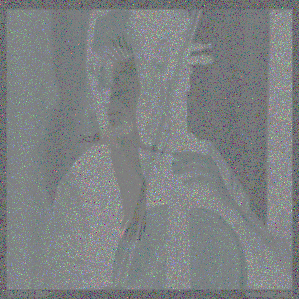} &
			\includegraphics[width=.195\textwidth]{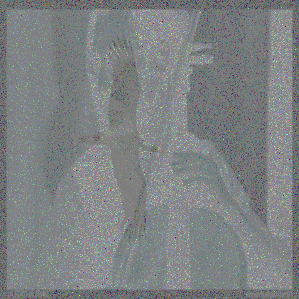} \\

			\\
			
			\textbf{Perlin bias only} & \textbf{Perlin+Surrogate} & \textbf{Perlin+Mask} & \textbf{Perlin+Mask+Surrogate} \\
			$d_{\ell^2} = 12.20$ & $d_{\ell^2} = 11.46$ & $d_{\ell^2} = 8.56$ & $d_{\ell^2} = 5.90$ \\
			\includegraphics[width=.195\textwidth]{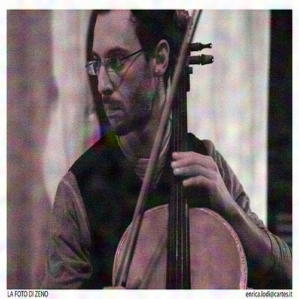} & 
			\includegraphics[width=.195\textwidth]{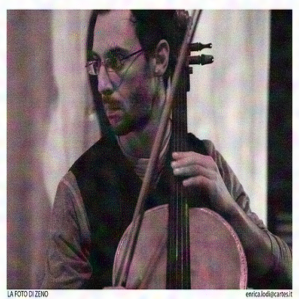} & 
			\includegraphics[width=.195\textwidth]{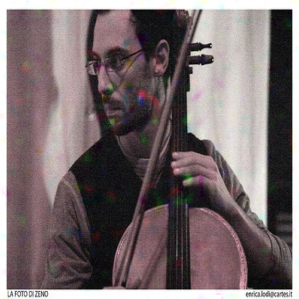} & 
			\includegraphics[width=.195\textwidth]{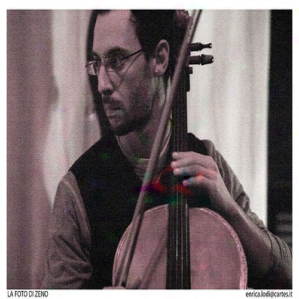} \\
			
			\includegraphics[width=.195\textwidth]{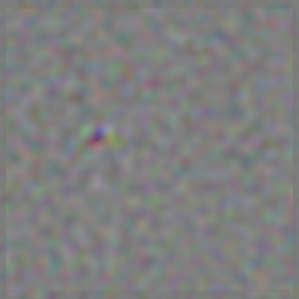} & 
			\includegraphics[width=.195\textwidth]{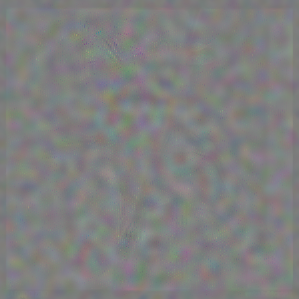} & 
			\includegraphics[width=.195\textwidth]{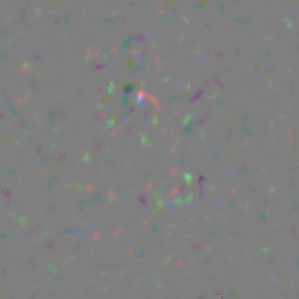} & 
			\includegraphics[width=.195\textwidth]{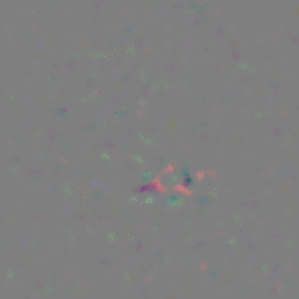} \\
			
			\\
			\textbf{Original image} & \textbf{Starting point} & & \\
			\includegraphics[width=.195\textwidth]{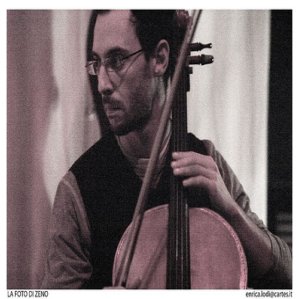} & 
			\includegraphics[width=.195\textwidth]{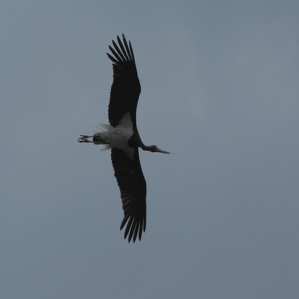} & 
			& 
			\\
			
		\end{tabular}	
		
		\label{fig-extra-cello}
	\end{center}
	\vskip -0.1in
	\caption{Targeted adversarial examples on ImageNet, obtained with different biases after 15000 iterations. The original class is "cello" -- all images are classified as the target "black stork". When used on its own, the surrogate bias seems to be detrimental for this particular image. Still, the final result is impressive: when comparing no biases with all biases, the perturbation norm is reduced by 88\%. See \url{https://github.com/ttbrunner/biased_boundary_attack} for an animated version.}
\end{figure*}

\clearpage
\section{Evaluation Hyperparameters}
\label{sec:appendix_hyper}

\subsection{Boundary Attack}

\noindent
\textbf{Step sizes}. In the source code of their original implementation, Brendel et al. \cite{brendel17} suggest setting both the orthogonal step to $\eta=0.01$ and the source step to $\epsilon=0.01$. Both step sizes are relative to the current distance from the source image. We instead set $\eta=0.05$ and $\epsilon=0.002$, as this makes the attack take smaller steps towards the source image, while at the same time allowing for more extreme perturbations in the orthogonal step. We have found this to increase the success chance of perturbation candidates, and the attack gets stuck less often.

\vskip 0.07in\noindent
\textbf{Step size adaptation}. We do not use the original step size adjustment scheme as proposed by Brendel et al. \cite{brendel17}. They collect statistics about the success of the orthogonal step before performing the step towards the source, and based on this they either reduce or increase the individual step sizes.

\vskip 0.07in\noindent
This method seems to be geared towards reaching near-zero perturbations and less towards query efficiency, which is our primary goal -- we are interested in making as much progress as possible in the early stages of an attack. When testing the Boundary Attack with query counts below 15000, we found the success statistics to be very noisy and the adaptation scheme ended up being detrimental. Therefore, we opt for a different approach:
\begin{itemize}
	\item At every iteration, we count the number of consecutive previously unsuccessful candidates.
	\item As this number increases, we dynamically reduce both step sizes towards zero.
	\item Whenever a perturbation is successful, the step size is reset to its original value.
	\item As a fail-safe, the step size is also reset after 50 consecutive failures. Typically, we found this to occur often for the unbiased Boundary Attack, but very seldom when using the Perlin bias.
\end{itemize}

\noindent
As a result, our strategy is quick to reduce step size, and after success immediately reverts to the original step size. We have found this to be very effective in the early stages of an attack. However, it has the drawback of wasting samples in the later stages (10000+ queries), when it tries to revert to larger step sizes too often. It might be promising to partially reinstate the approach of Brendel et al., or to apply some form of step size annealing.

\subsection{Other attacks}

\noindent
For all other attacks, we use the hyperparameters that are provided for ImageNet in the publicly available source code of their implementations.

\clearpage
\section{Submission to NeurIPS 2018 Adversarial Vision Challenge}
\label{sec:appendix_avc}

\noindent
When evaluating adversarial attacks and defenses, it is hard to obtain meaningful results. Very often, attacks are tested against weak defenses and vice versa, and results are cherry-picked. We sidestep this problem by instead presenting our submission to the NeurIPS 2018 Adversarial Vision Challenge (AVC), where our method was pitted against state-of-the-art robust models and defenses and won second place in the targeted attack track.

\vskip 0.07in\noindent
\textbf{Evaluation setting}. The AVC is an open competition between image classifiers and adversarial attacks in an iterative black-box decision-based setting \cite{avc2018}. Participants can choose between three tracks:
\begin{itemize}
	\item Robust model: The submitted code is a robust image classifier. The goal is to maximize the $\ell^2$ norm of any successful adversarial perturbation.
	\item Untargeted attack: The submitted code must find a perturbation that changes classifier output, while minimizing the $\ell^2$ distance to the original image.
	\item Targeted attack: Same as above, but the classification must be changed to a specific label.
\end{itemize}

Attacks are continuously evaluated against the current top-5 robust models and vice versa. Each evaluation run consists of 200 images with a resolution of 64x64, and the attacker is allowed to query the model 1000 times for each image. The final attack score is then determined by the median $\ell^2$ norm of the perturbation over all 200 images and top-5 models (lower is better).

\vskip 0.07in\noindent
\textbf{Competitors}. At the time of writing, the exact methods of most model submissions were not yet published. But seeing as more than 60 teams competed in the challenge, it is reasonable to assume that the top-5 models accurately depicted the state of the art in adversarial robustness. We know from personal correspondence that most winning models used variations of Ensemble Adversarial Training \cite{tramer17ens}, while denoisers were notably absent. On the attack side, most winners used variants of PGD transfer attacks, again in combination with large adversarially-trained ensembles.

\vskip 0.07in\noindent
\textbf{Dataset}. The models are trained with the Tiny ImageNet dataset, which is a down-scaled version of the ImageNet classification dataset, limited to 200 classes with 500 images each. Model input consists of color images with 64x64 pixels, and the output is one of 200 labels. The evaluation is conducted with a secret hold-out set of images, which is not contained in the original dataset and unknown to participants of the challenge.

\subsection{Random guessing with low frequency}

\noindent
Before implementing the biased Boundary Attack, we first conduct a simple experiment to demonstrate the effectiveness of Perlin noise patterns against strong defenses. Specifically, we run a random-guessing attack that samples candidates uniformly from the surface of a $\ell^2$-hypersphere with radius $\epsilon$ around the original image:

\vskip -0.05in
\begin{equation}
s \sim \mathcal{N}(0, 1)^k;\ x_{adv} = x_0 + \epsilon \cdot \frac{s}{\|s\|_2}
\end{equation}

\noindent
With a total budget of 1000 queries to the model for each image, we use binary search to reduce the sampling distance $\epsilon$ whenever an adversarial example is found. First experiments have indicated that the targeted setting may be too difficult for pure random guessing. Therefore we limit this experiment to the untargeted attack track, where the probability of randomly sampling \textit{any  of 199} adversarial labels is reasonably high. We then replace the distribution with normalized Perlin noise:

\vskip -0.05in
\begin{equation}
s \sim Perlin_{64,64}(v);\ x_{adv} = x_0 + \epsilon \cdot \frac{s}{\|s\|_2}
\end{equation}

\noindent
We set the Perlin frequency to 5 for all attacks on Tiny ImageNet. As Table \ref{tab1} shows, Perlin patterns are more efficient and the attack finds adversarial perturbations with much lower distance (63\% reduction). Although intended as a dummy submission to the AVC, this attack was already strong enough for a top-10 placement in the untargeted track. An example obtained in this experiment can be seen in Figure \ref{fig:figure-perlin}.

\begin{table}[htbp]
	\parbox{.45\linewidth}{

		\vskip 0.1in
		\begin{center}
			\begin{small}
				\begin{sc}
					\begin{tabular}{lcr}
						Distribution & \ \ \ \ \ \ \ \ \ \ \ \ \ \ \  & Median $\ell^2$ \\
						\hline
						Normal 		&	& 11.15 \\
						\textbf{Perlin noise}&	& \textbf{4.28} \\
					\end{tabular}
				\end{sc}
			\end{small}
		\end{center}
		\vskip 0.165in
		\caption{Random guessing with low frequency (untargeted), evaluated against the top-5 models in the AVC.}
				\label{tab1}
	}\ \ \ \ \ \ \ 
	\parbox{.45\linewidth}{

		\vskip 0.15in
		\begin{center}
			\begin{small}
				\begin{sc}
					\begin{tabular}{lr}
						Boundary Attack Bias & Median $\ell^2$  \\
						\hline
						None 			& 20.2 \\
						Perlin		& 15.1 \\
						\textbf{Perlin + Surrogate gradients}	& \textbf{9.5} \\
					\end{tabular}
				\end{sc}
			\end{small}
		\end{center}
		\caption{Biases for the Boundary Attack (targeted), evaluated against the top-5 models in the AVC.}
				\label{tab2}
	}
	\vskip 0.1in

\end{table}

\subsection{Biased Boundary Attack}

\begin{figure}[htb]
	\begin{center}
		\begin{tabular}{c c c c}
			
	\textbf{Original} & \textbf{Unbiased} & \textbf{Perlin} & \textbf{Perlin + Surrogate} \\
	\includegraphics[width=.21\textwidth]{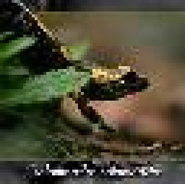} &
	\includegraphics[width=.21\textwidth]{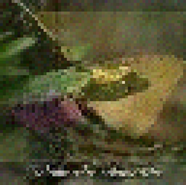} &
	\includegraphics[width=.21\textwidth]{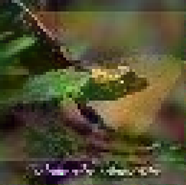} & 
	\includegraphics[width=.21\textwidth]{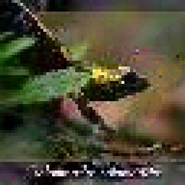} \\
	European fire salamander & Sulphur butterfly & Sulphur butterfly & Sulphur butterfly \\		
	\\
	\includegraphics[width=.21\textwidth]{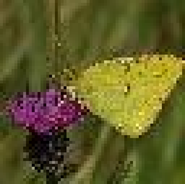} &
	
	\includegraphics[width=.21\textwidth]{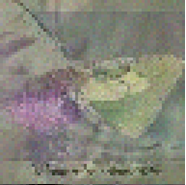} &
	\includegraphics[width=.21\textwidth]{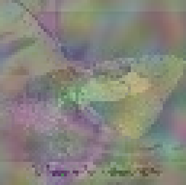} & 
	\includegraphics[width=.21\textwidth]{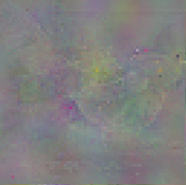} \\
	Starting image & $d_{\ell^2} = 9.4$ & $d_{\ell^2} = 7.5$ & $d_{\ell^2} = 4.4$ \\		
		\end{tabular}	
	\end{center}
	\caption{Adversarial examples generated with different biases in our targeted attack submission to the AVC. All images were obtained after 1000 queries. The isolated perturbation is shown below each adversarial example.}
	\label{fig:salamander-bba}	
	\vskip -0.1in
\end{figure}

\noindent
Next, we evaluate the biased Boundary Attack in our intended setting, the targeted attack track in the AVC. 
To provide a point of reference, we first implement the original Boundary Attack without biases. This works, but is too slow for our setting. Compare Figure \ref{fig:salamander-bba}, where the starting point is still clearly visible after 1000 iterations (in the unbiased case).

\noindent
\textbf{Perlin bias}. We add our first bias, low-frequency noise. As before, we simply replace the distribution from which the attack samples the orthogonal step with Perlin patterns. See Table \ref{tab2}, where this alone decreases the median $\ell^2$ distance by 25\%.  

\vskip 0.07in\noindent
\textbf{Surrogate gradient bias.} We also add projected gradients from a surrogate model and set the bias strength $w$ to 0.5. This further reduces the median $\ell^2$ distance by another 37\%, or a total of 53\% when compared with the original Boundary Attack. 1000 iterations are enough to make the butterfly almost invisible to the human eye (see Figure \ref{fig:salamander-bba}).

\vskip 0.07in\noindent
Here, the efficiency boost is much larger than in our ImageNet evaluation in Section \ref{sec:eval}. This may be due to our choice of surrogate models: In our submission to the AVC, we simply combined the publicly available baselines (ResNet18 and ResNet50). This ensemble is notably stronger than the simple model we used for the ImageNet evaluation, as the ResNet50 model is adversarially trained. However, it is also significantly weaker than the ones used by other winning AVC attack submissions, most of which were found to use much larger ensembles of carefully-trained models.\footnote{\url{https://medium.com/bethgelab/results-of-the-nips-adversarial-vision-challenge-2018-e1e21b690149}} Nevertheless, our attack outperformed most of them which reinforces our earlier claim: Our method seems to make more efficient use of surrogate models than direct transfer attacks.

\vskip 0.07in\noindent
\textbf{Mask Bias}. We did not implement the mask bias in our entry to the AVC because of time constraints.

\vskip 0.07in\noindent
The source code of our submission is publicly available.\footnote{\url{https://github.com/ttbrunner/biased_boundary_attack_avc}}

\end{document}